\documentclass[letterpaper, 10 pt, conference]{ieeeconf}
\IEEEoverridecommandlockouts                              
\usepackage{graphicx,dblfloatfix}
\usepackage{amsmath} 
\usepackage{amssymb}  
\usepackage{subcaption}
\usepackage{array}
\usepackage{cite}   
\usepackage{hyperref}
\usepackage{float}
\usepackage{mathtools}

\hypersetup{
    colorlinks=true,
    linkcolor=black,
    urlcolor=black,
}
\include{notation}

\def\showcomments{1}
\usepackage{xcolor}
\newcommand{\peter}[1]{{
    \if\showcomments1
        \color{red}PIC: #1
    \fi
}}

\newcommand{\juxicomment}[1]{{
    \if\showcomments1
        \color{red}JL: #1
    \fi
}}

\title{\LARGE \bf DGBench: An Open-Source, Reproducible Benchmark for Dynamic Grasping}
\author{Ben Burgess-Limerick$^{1}$, Chris Lehnert$^{1}$,  J\"urgen Leitner$^{2}$, Peter Corke$^{1}$
\thanks{This research was supported by the QUT Centre for Robotics.}
\thanks{$^{1}$Ben Burgess-Limerick, Chris Lehnert, and Peter Corke are with the Queensland University of Technology Centre for Robotics (QCR), Brisbane, Australia
        {\tt\small ben.burgesslimerick@qut.edu.au}}
\thanks{$^{2}$J\"urgen Leitner is with LYRO Robotics, Brisbane, Australia}%
}

\begin{document}
\maketitle
\thispagestyle{empty}
\pagestyle{empty}
\begin{abstract}
This paper introduces DGBench, a fully reproducible open-source testing system to enable benchmarking of dynamic grasping in environments with unpredictable relative motion between robot and object. We use the proposed benchmark to compare several visual perception arrangements. Traditional perception systems developed for static grasping are unable to provide feedback during the final phase of a grasp due to sensor minimum range, occlusion, and a limited field of view. A multi-camera eye-in-hand perception system is presented that has advantages over commonly used camera configurations. We quantitatively evaluate the performance on a real robot with an image-based visual servoing grasp controller and show a significantly improved success rate on a dynamic grasping task.

\end{abstract}


\setcounter{footnote}{2}
\section{Introduction}

The ability to quickly and reliably grasp objects in arbitrary environments is a key capability for many robot applications. In real-world tasks it is often necessary to interact with dynamic objects or interact with objects while a robot is in motion. Both of these scenarios present the opportunity for relative motion between hand and object that increases the difficulty of reliably performing grasps. For example, grasping an object while a mobile manipulator is in motion, or from an underwater or aerial robot, requires the system to react to relative motion between the robot and object. Tasks where the object is in motion, such as human-to-robot hand overs or grasping in the zero-gravity environment of space, present the same challenges.

A number of approaches to evaluating grasp performance have been reported \cite{MahlerBenchmark}, but dynamic grasping performance is largely anecdotal and there are no standardised methods. Here we present DGBench, a physical dynamic grasping benchmark that enables reproduction and comparison of system performance. In this work, we use the proposed benchmark to compare several perception system designs.

Modern closed-loop systems revert to open-loop control during the final phase of grasping, up to object contact, which admits error in the case of relative motion. 
The open-loop operation is due to limitations of perception such as sensor minimum range, limited camera field of view, or occlusions by clutter.
One approach to overcoming this is to estimate and predict the relative motion, however this requires an assumed motion model. 
Alternatively, the perception system can be designed so that we observe the grasp until the fingers close on the object.

\begin{figure}[t]
\begin{subfigure}{\linewidth}
\centering
\includegraphics[width=\linewidth]{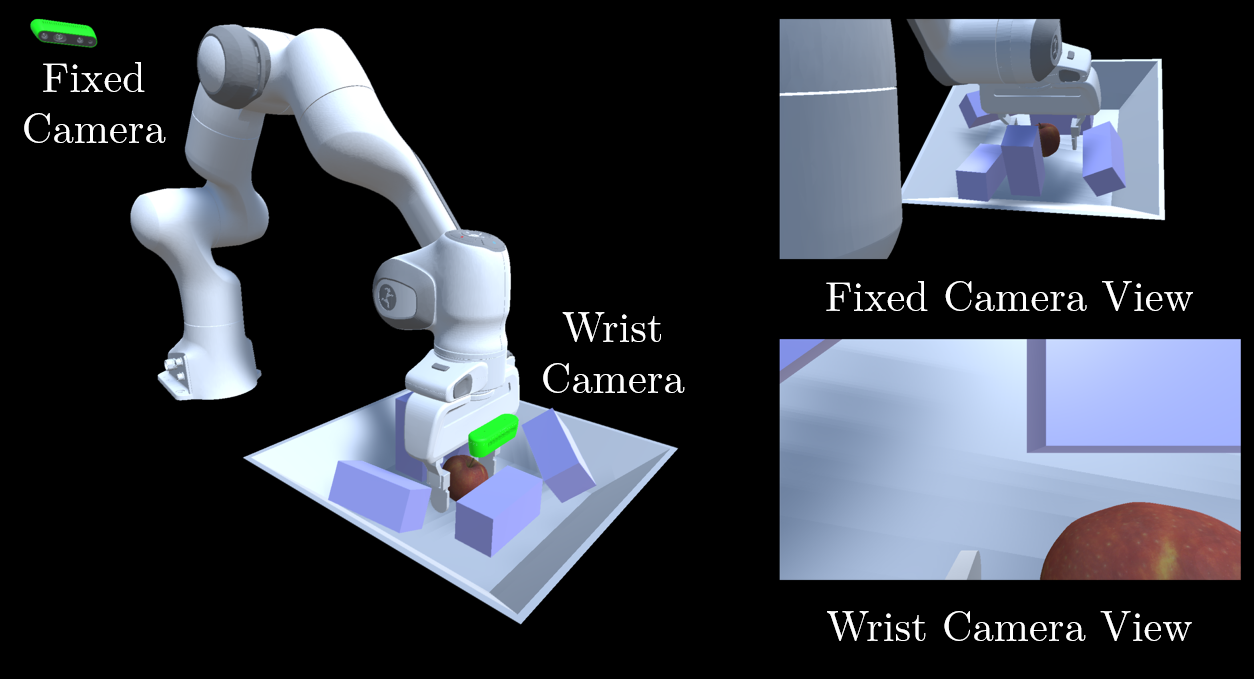}
\caption{Typical fixed and wrist-mounted camera configurations}
\label{fig:WristShoulder}
\end{subfigure}

\begin{subfigure}{\linewidth}
\vspace{2mm}
\centering
\includegraphics[width=\linewidth]{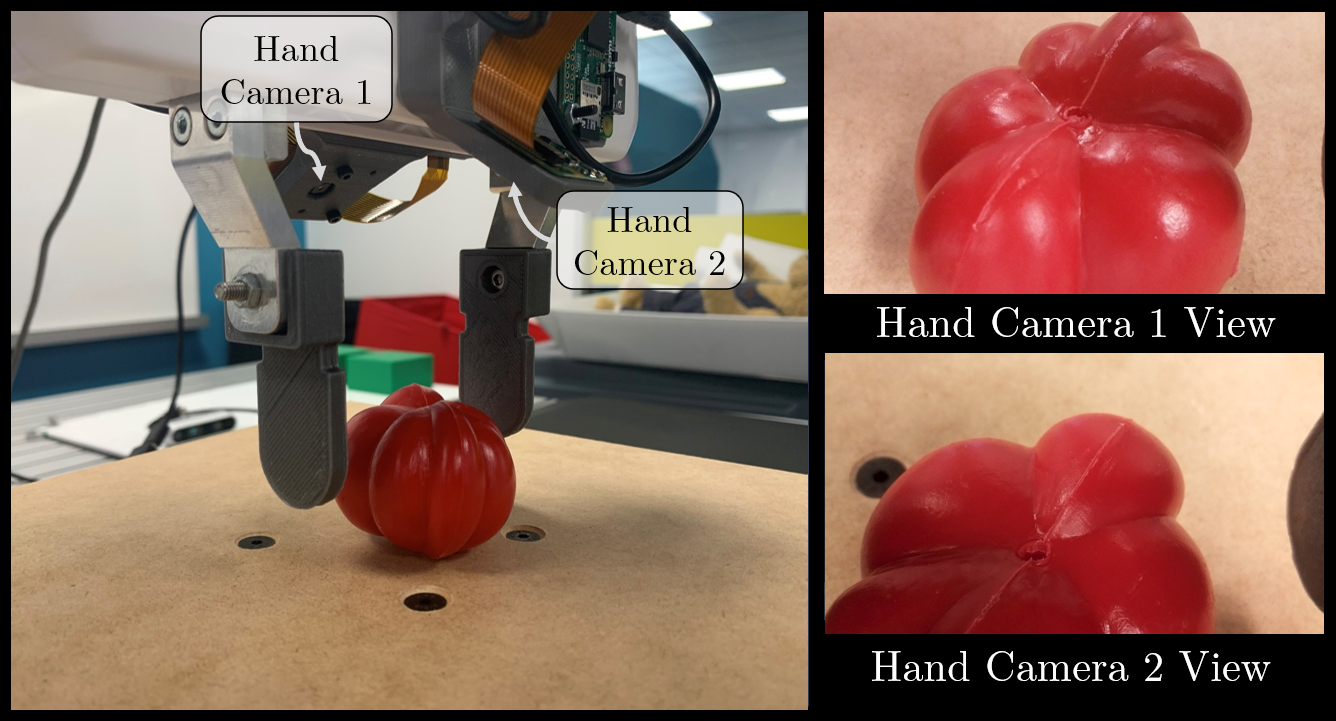}
\caption{Proposed multi-camera configuration}
\label{fig:HandCams}
\end{subfigure}
\caption{Camera placement is critical to ensuring continuous observation during the final phase of grasping, particularly in dynamic scenarios.}
\vspace{-0.3cm}
\end{figure}

Fig. \ref{fig:WristShoulder} illustrates the view from two conventional camera placements during the final phase of grasping from a cluttered scene. Due to occlusion and limited field of view, these cameras provide limited feedback on the object and may lose perception in the event of unexpected motion. Furthermore, when close to the object, common wrist-mounted depth cameras fail due to sensor minimum range.

In contrast, Fig. \ref{fig:HandCams} presents a grasp-perception system that uses two cameras directed between the robot's fingers to provide uninterrupted feedback throughout the grasping action. 

The principal contributions of this work are:
\begin{enumerate}
    \item DGBench, a performance benchmark for dynamic grasping that includes an open-source physical testbed and performance measures\footnote{\url{https://github.com/BenBurgessLimerick/DGBench}}. 
    \item A perception system designed for closed-loop visual feedback for the \textit{entire} grasp -- until the fingers contact the object.
     \item A comparative assessment of robotic grasping performance for common camera positions and the proposed arrangement.

\end{enumerate}

\section{Related Works}
\subsection{Benchmarking for Robotic Grasping}
Previous works have explored a range of different dynamic grasping tasks. Some works demonstrate successful re-grasping after an initial grasp attempt causes the object to move \cite{Kalashnikov}. Others explore objects moving along predictable trajectories at constant speeds \cite{Allen, Akinola}. A third common task is grasping an object that is moved by a human, however the motion is stopped before the robot's fingers close on the object \cite{Viereck, Tuscher, MorrisonLearningRobust, Song}. Dynamic grasping of objects with unpredictable trajectories and accelerations is demonstrated in simulation in \cite{Akinola}. However, in this case, the position of the object is observed directly from the simulation, without a perception system. 

These various definitions of dynamic grasping highlight the need for standardisation of dynamic grasping tasks in order to compare proposed solutions. Reproducible benchmarking for robotic grasping is a critical step for advancing the field \cite{MahlerBenchmark, Bonsignorio}. Early evaluations of static grasping systems used an array of different objects and tasks that made direct comparisons of performance impossible \cite{Calli}. The introduction of standardised objects sets and testing methods has significantly reduced the complexity of comparing approaches.

To the best of the authors' knowledge there is no standardised benchmark for evaluating dynamic grasping performance. \cite{MahlerBenchmark} highlights the importance of presenting repeatable benchmarks with clear definitions of the hardware setup, testing method, and performance metrics. In this work we present a benchmark for dynamic grasping in the form of open-source hardware and software for a standardised dynamic workspace. This is combined with example results comparing several perception systems on a dynamic grasping task. 

\subsection{Grasping Perception Systems}
Robotic grasping has a long history and numerous perception systems have been explored. Many different analytic and data-driven grasp synthesis methods have been proposed, and have been summarised in several surveys \cite{Shimoga, Bicchi, Sahbani, Bohg, Kleeberger}. This review focuses on the strengths and limitations of the perception systems used in prior work rather than the grasp synthesis problem itself. In particular, focus will be given to how these perception systems perform in dynamic grasping tasks. 

Some grasping systems use tactile sensing to explore and refine grasp points \cite{Chebotar, Yu, Calandra, Farias}. However, tactile sensors are unable to provide feedback until they come into contact with the object and are not well suited to dynamic environments where object motion cannot be observed when the fingers are not in contact. We review only visual perception systems here. 

\subsubsection{Fixed Cameras}
Early robotic grasping systems utilised cameras that were fixed in the world frame to identify grasp points \cite{Hutchinson, Bicchi, Kamon}. This can be a robust solution for tasks in a structured environment. The camera can be positioned such that it is always looking at the target objects and its view is not occluded by the robot. Utilising multiple cameras can help mitigate against occlusion caused by clutter in the scene. Typically, cameras are mounted on the opposite side of the workspace from the robot in order to reduce occlusion. Several works explore methods for grasping in the presence of occlusion \cite{Lin, Yu}.

In the context of dynamic grasping, fixed cameras can provide consistent feedback on object motion in the early stages of grasping provided object motion is limited to the camera's fixed field of view. However, as the robot's fingers approach the object they may occlude the camera's view \cite{Tuscher}. Under these conditions, a robust software solution is required to maintain tracking when the target object is partially occluded \cite{Tuscher}. \cite{Allen} presents a system that uses two fixed cameras to successfully grasp a toy train moving along an oval trajectory. 

More recently, fixed perception systems have been used to further explore the planning difficulties associated with grasping a moving object \cite{Marturi, Akinola}. The tasks demonstrated include a human operator moving the object by hand \cite{Tuscher, Marturi}, objects moving on a straight conveyor \cite{Akinola}, and objects moving on more complex trajectories \cite{Akinola}. The more complex trajectories presented in \cite{Akinola} are demonstrated in simulation only, without a perception system. 

For structured dynamic grasping tasks, such as picking from a conveyor, multiple fixed cameras provide sufficient feedback for reliable grasping, despite occlusion challenges. 

\subsubsection{Shoulder-Mounted Cameras}
\label{section:ShoulderMounted}
For a mobile robot there are constraints on where cameras can be located. Cameras that remain fixed in robot frame are usually mounted in a shoulder or head position. These systems have similar advantages to general fixed cameras but are faced with increased self-occlusion where the manipulator obstructs the view of the object. Several works have been presented with cameras in this configuration \cite{Pinto, Kalashnikov, Levine}. 

Self-occlusion can be reduced by controlling the manipulator to avoid the line of sight to the object \cite{Logothetis}. However this places significant constraints on the manipulator motion, particularly when the hand is close to the object. The system presented in \cite{Logothetis} disables self-occlusion avoidance when the hand is very close to the object, resulting in significant occlusion during the final phase. In a dynamic environment the loss of perception due to occlusion may cause grasp failure.

\subsubsection{Wrist-Mounted Cameras}
Mounting a camera on the wrist of a robot can significantly reduce the impact of occlusion on a grasping task. If the system does not have a good view of the object it can be moved so that it does \cite{Kahn, MorrisonMultiView}. Most grasping systems of this type utilise an RGB-D camera which provides depth information that can be used to identify grasp points \cite{Viereck, MorrisonLearningRobust, Song, Mousavian, ZengTossingBot, Haviland, Berscheid, Johns}. This can be a robust solution for static grasping tasks. However, depth cameras cannot provide useful data when objects are closer than their minimum range which is often the case in the final stages of a grasp \cite{MorrisonLearningRobust}. This presents a significant challenge for dynamic grasping because closed-loop systems must perform the final phase of a grasp open-loop due to the sensor minimum range. The grasp attempt will fail should unexpected motion occur at this time.

Several works present closed-loop systems using a wrist-mounted depth camera that demonstrate grasping in a dynamic environment by moving the object during the robot's approach. However, the motion is stopped before the final phase of grasping when the sensor's minimum range is violated \cite{Viereck, MorrisonLearningRobust, Song}.

Using an RGB camera without depth removes the minimum range limitation of depth cameras. A wrist-mounted RGB camera can provide reliable feedback for a dynamic grasping system. However, when the hand is close to the object, the camera's perception is limited by its field of view. Typically, wrist-mounted cameras are positioned parallel to the final robot axis and directed to look above the robot's fingers \cite{Pinto, Saxena}. This is a convenient and common mounting location and helps to limit the occlusion caused by the robot's fingers during the initial grasp point identification. However, when the robot is about to grasp the object, the object is typically on the border of the camera's field of view. In a dynamic environment, a sudden motion of the object can move the object out of the camera's view resulting in a loss of perception and potentially a failed grasp \cite{Haviland}. 

\cite{Haviland} presents a system that expands on that of \cite{MorrisonLearningRobust} by switching from depth-based control to RGB-based control when the robot enters the final phase of grasping. A single RGB-D sensor is used that is mounted parallel to, and offset from, the final robot axis. The authors note that the system can only handle relatively small motions of the object because large motions move the object outside the field of view of the camera and tracking is lost \cite{Haviland}. Therefore, \cite{Haviland} demonstrates grasping on tasks involving relatively small object movements, and where the motion stops before the grasping action is completed. 

As noted in Section \ref{section:ShoulderMounted}, mobile robotics applications limit where a fixed camera can be mounted and wrist-mounted sensors are a popular choice. \cite{Arora} uses a wrist-mounted sensor to observe and grasp an object that is moved around by hand. \cite{Arora} simplifies the perception challenge by affixing an Apriltag marker to the object, and instead focuses on the visual servoing challenge presented by a dynamic object and mobile base. \cite{Zimmermann} presents a system with a Kinova manipulator mounted to a Boston Dynamics Spot. The system uses a wrist-mounted camera to demonstrate tasks such as grasping a ball from the floor as the quadruped walks past. However, the authors note that they execute the trajectory in an open-loop fashion after initial identification of the ball position. Consequently the system is prone to failure resulting from unexpected motion of the mobile base and inaccurate object pose estimation. The system presented in \cite{Zimmermann} highlights the need for closed-loop dynamic grasping systems in mobile robotics applications.

\section{Method}

\subsection{DGBench Dynamic Workspace Hardware}
\begin{figure}[t]
\vspace{6pt}
\centerline{\includegraphics[width=\linewidth]{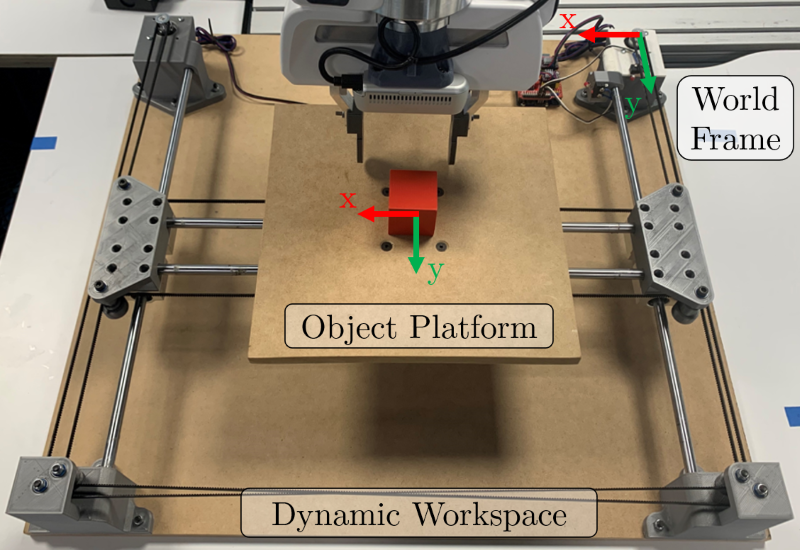}}
\caption{DGBench dynamic workspace system. The object platform is controlled to induce arbitrary yet repeatable motion of the object relative to the robot.}
\label{fig:DynamicWorkspace}
\end{figure}

The design shown in Fig. \ref{fig:DynamicWorkspace} presents an XY controllable platform that is driven by two stepper motors and controlled by an Arduino running Grbl\footnote{\url{https://github.com/gnea/grbl}}. The object platform measures 260$\times$260 mm and can move at speeds up to 250 mm/s within a 280$\times$280 mm area. The platform is controlled with G-code commands that give precise control of the induced object motion. 

\subsection{DGBench Trajectory Design}
The dynamic object trajectories implemented in this work are designed to produce motion that is unpredictable to the grasp system and requires continuous observation of the object for a successful grasp to be performed. When an object moves at high speed and accelerates frequently it is difficult to predict the motion and calculate an intercepting grasp point. In our experiment, the frequency of direction change and absolute object speed are varied together by moving the object along a given trajectory at various speeds. 

\begin{figure}[t]
\centering
\includegraphics[width=\linewidth]{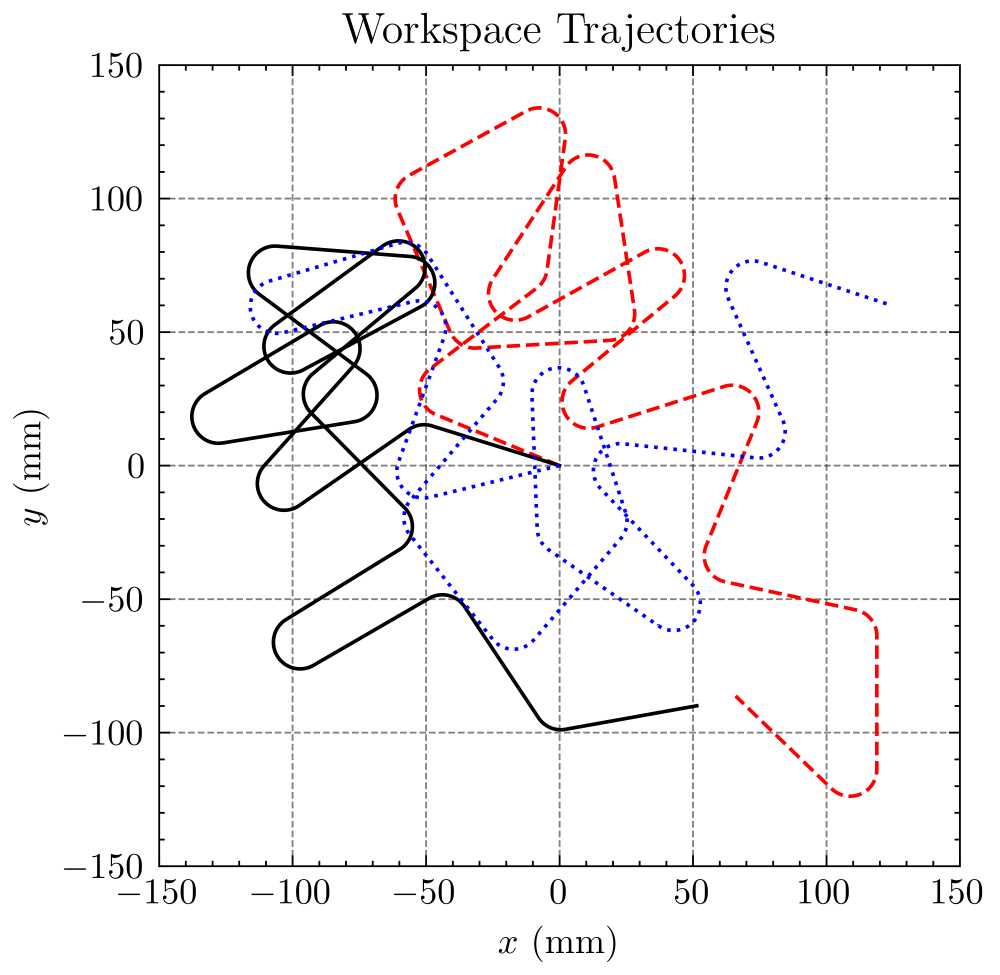}
\caption{Three example trajectories generated with direction changes every 50 mm and 10 mm radius corners. These paths are executed on the platform presented in Fig. \ref{fig:DynamicWorkspace}.}
\label{fig:RandomWalkTrajectories}
\end{figure}

Trajectories are generated by arranging linear segments of a constant length connected with random direction changes. All trajectories start in the centre of the workspace and proceed randomly around the area. Direction changes are of random magnitude between 45\textdegree\ and 315\textdegree. The direction changes are smoothed by joining each segment with an arc of fixed radius. The trajectories were generated with linear segments of length 50 mm and 10 mm radius fillets. Fig. \ref{fig:RandomWalkTrajectories} shows the first 15 segments of 3 generated trajectories.

For this work, a family of 20 trajectories was generated, and the performance of each grasping system was evaluated on each trajectory at speeds from 100 mm/s to 200 mm/s.  

\subsection{DGBench Trial Procedure}
Each individual trial was conducted according to the following procedure. The result of the grasp attempt was recorded along with the time between the start of step 2 and end of step 3:

\begin{enumerate}
    \item Robot drives to a home position 500 mm above the workspace and fingers are opened.
    \item Simultaneously, the visual servoing controller is enabled and the dynamic workspace begins moving along a randomly chosen trajectory.
    \item The robot follows the actions of the controller until its fingers are closed.
    \item Once fingers are closed, the object is lifted from the workspace.
    \item Grasp success is evaluated and the experiment is reset.
\end{enumerate}

Each trial ends in one of three outcomes. The grasp attempt is successful if the object is lifted from the platform by the gripper. If object motion removes the object from the camera's field of view, the attempt is abandoned and a perception failure is recorded. If observation is maintained throughout the attempt but contact from the fingers causes the object to fall from the platform, or the object escapes the closing gripper, a grasp failure is recorded.

Object size is a crucial parameter for evaluating the effectiveness of a dynamic grasping perception system, so each trial was evaluated with 3D printed red cubes of side length 30 mm and 40 mm. The simple object geometry removes the need for a grasp point selection method. The cube starts each trial in the centre of the platform, with its axes aligned to the workspace.

The proposed task is relatively easy for a human to complete due to our high-speed motion capability relative to the tested object speeds, and our ability to use multiple fingers in a power grasp configuration to restrict object motion. However, for a manipulator with a lower maximum speed and equipped with an antipodal gripper the task presents a significant challenge.

\subsection{Proposed Perception System}
\label{section:percetion_system}

\begin{figure}[t]
\vspace{6pt}
\begin{subfigure}{\linewidth}
\centering
\includegraphics[width=\linewidth]{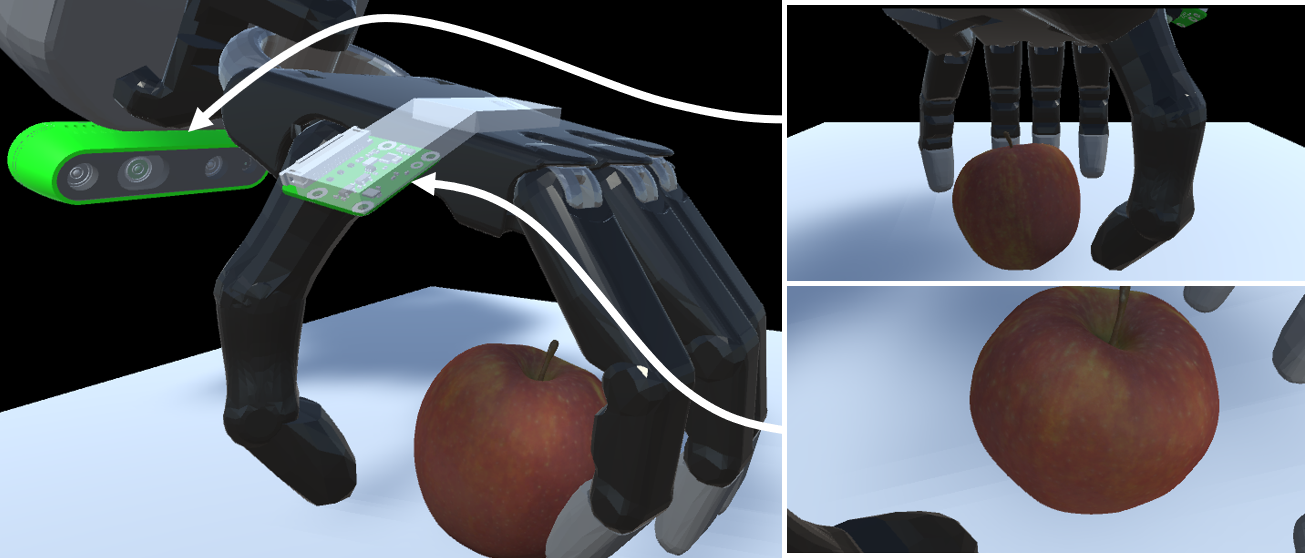}
\caption{Hand and wrist camera system on a Shadow Dexterous Hand}
\label{fig:ShadowPerception}
\end{subfigure}
\begin{subfigure}{\linewidth}
\centering
\includegraphics[width=\linewidth]{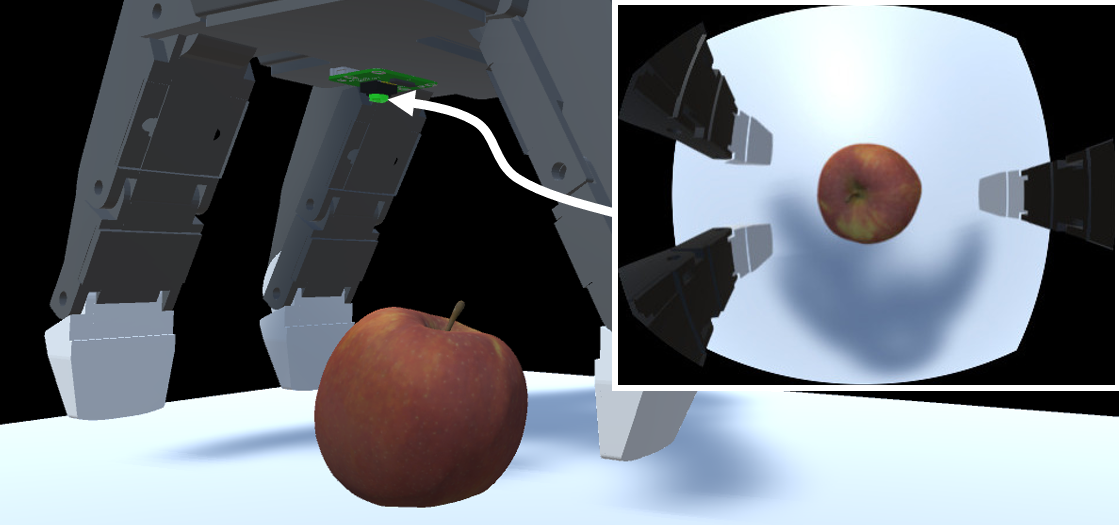}
\caption{Single fish-eye palm camera on a Robotiq 3 Finger gripper.}
\label{fig:RobotiqPerception}
\end{subfigure}
\caption{Final phase perception systems for complex grippers.}
\label{fig:PerceptionSystems}
\end{figure}

\begin{figure*}[t]
\vspace{10pt}
\begin{subfigure}{.5\textwidth}
\centering
\includegraphics[width=\linewidth]{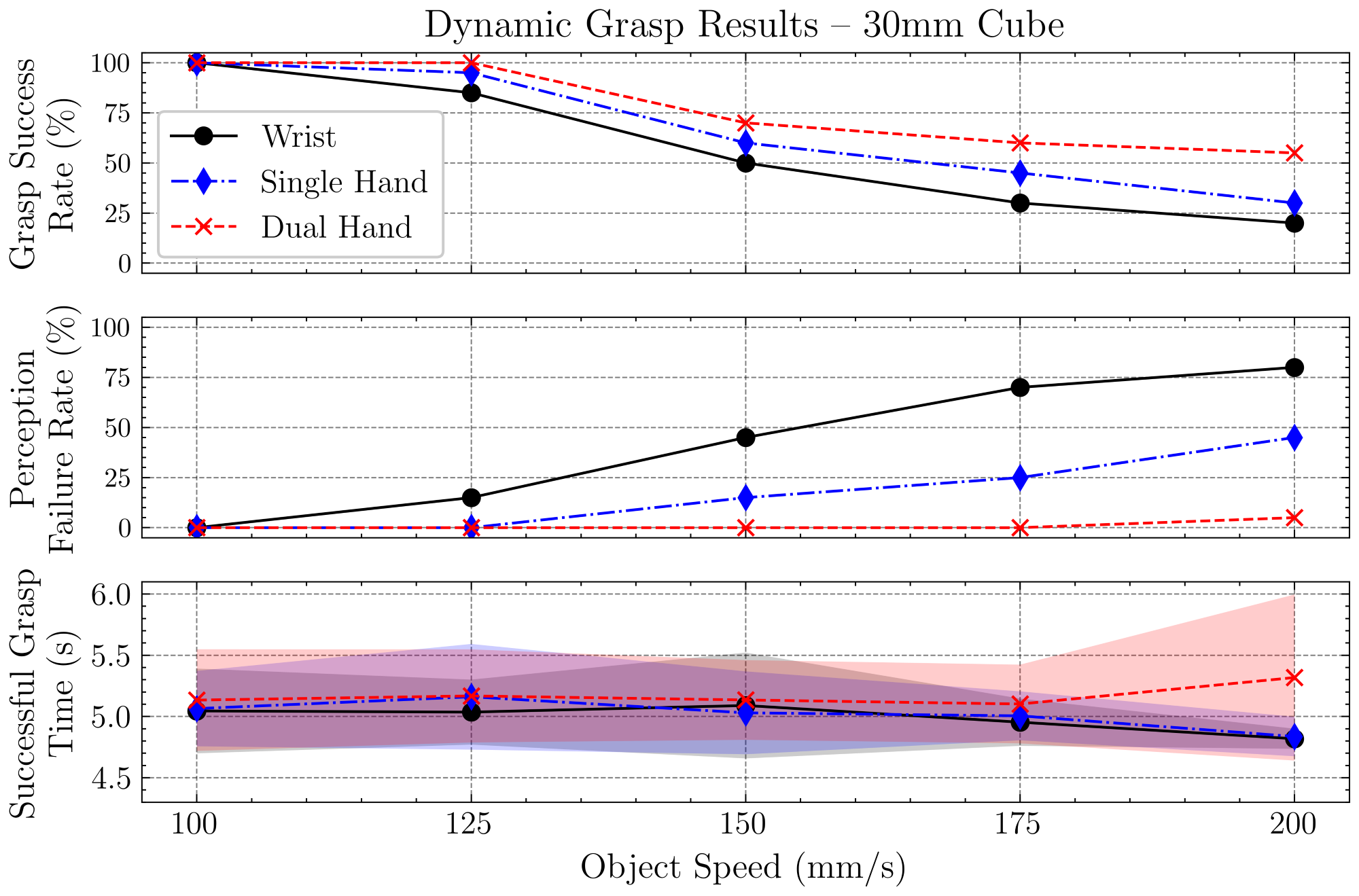}
\caption{Dynamic grasping results with 30mm cube.}
\label{fig:30mmCubeResults}
\end{subfigure}
\begin{subfigure}{.5\textwidth}
\centering
\includegraphics[width=\linewidth]{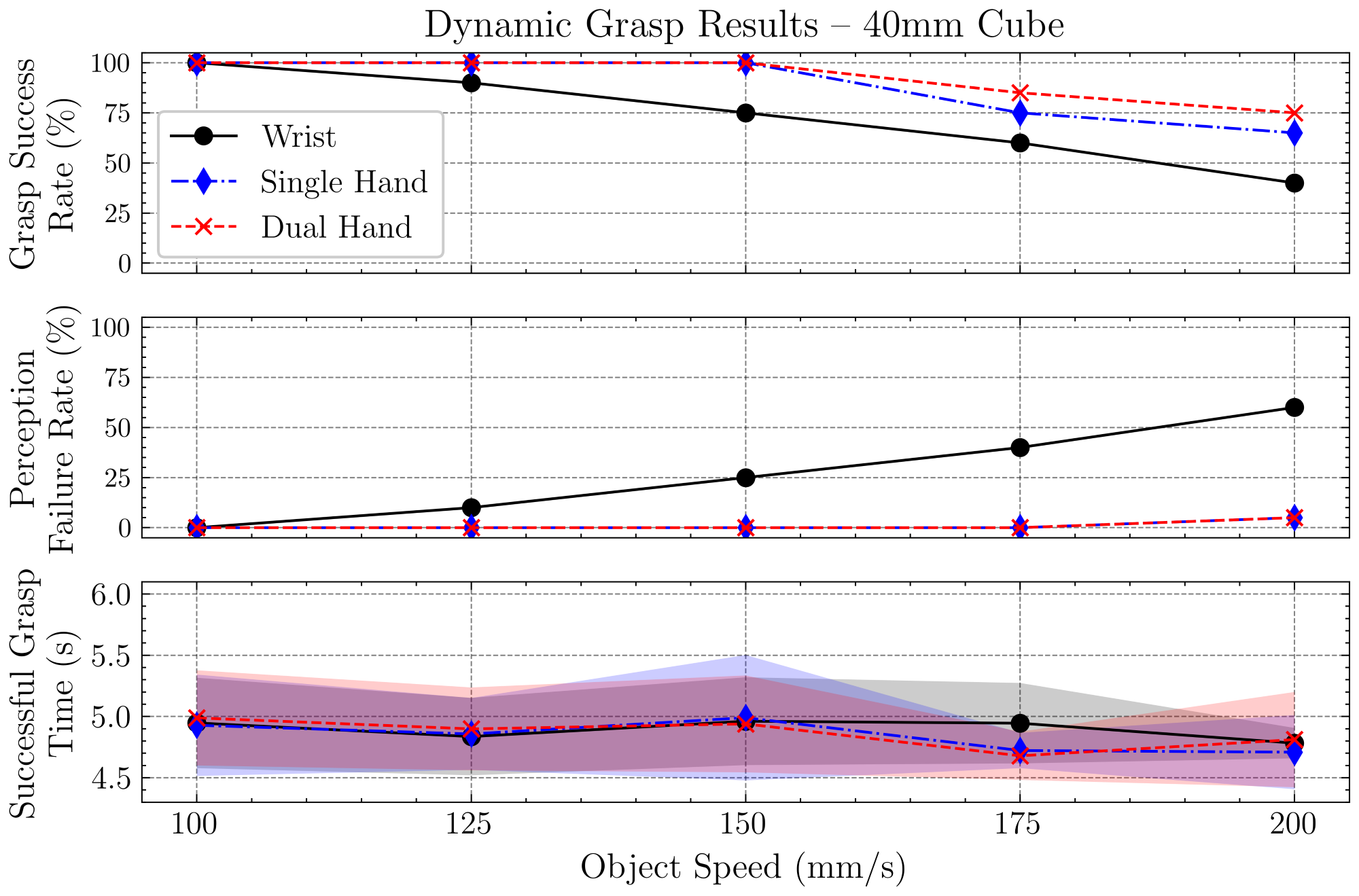}
\caption{Dynamic grasping results with 40mm cube.}
\label{fig:40mmCubeResults}
\end{subfigure}
\caption{Grasp success rates, perception failure rates, and grasp times across object speeds, cube sizes, and perception systems. The shaded areas on successful grasp times show $\pm1\sigma$. The hand-mounted camera configurations demonstrate improved performance compared to the conventional wrist-mounted system.}
\label{fig:GraspingResults}
\end{figure*}

We investigate the performance benefits for dynamic grasping of a perception system designed to overcome occlusion, limited field of view, and sensor minimum range compared to a traditional camera-on-wrist configuration. Fixed perception systems either mount the camera externally to the robot, or experience significant self-occlusion in the final phase of a grasp. External mounting is not suitable for common dynamic grasping environments such as mobile manipulation, and self-occlusion results in consistent failure when objects are moving quickly and unpredictably. Therefore we limit our comparison to wrist-mounted cameras. 

Fig. \ref{fig:HandCams} presents the multi-camera eye-in-hand perception system investigated. The camera positions are selected to minimise occlusion in the final phase of a grasp. Multiple cameras are included to increase the system's field of view. The same guiding design principles can be used to implement final phase perception systems for more complex grippers. Fig. \ref{fig:ShadowPerception} presents a perception system for an anthropomorphic hand which avoids self-occlusion and improves field of view by moving the wrist camera below the wrist and including a dedicated hand camera. Similarly, a single fish-eye camera in the palm of a 3 finger gripper can achieve the same goals (Fig. \ref{fig:RobotiqPerception}). 

In this work we consider an antipodal gripper performing a top-down grasp on a moving object and compare three different perception systems:
\begin{enumerate}
    \item Conventional wrist-mounted camera
    \item Single-hand camera pointed between fingers
    \item Dual-hand cameras pointed between fingers 
\end{enumerate}

\subsection{Visual Servoing Controller}
The manipulator is controlled by an image-based visual servoing control system \cite{Hutchinson}. The object centroid is identified using image segmentation and the robot is controlled such that the identified centroid is driven towards a predetermined position in image space. This goal position is predetermined by calculating the position of the object in image space when it is centred between the robot's fingers. The robot is controlled in \(x\) and \(y\) parallel to the workspace while descending at a constant rate. The orientation of the gripper is constant. These degrees of freedom are sufficient to successfully grasp a simple object.

The following procedure presents the control system for calculating a robot action from an input image. Each robot action includes a desired end-effector velocity vector and a binary close fingers command. This procedure occurs for every input frame:

\begin{enumerate}
    \item Receive new RGB image (25 Hz in our case)
    \item Segment pixels (red $>$ blue + green + 50)
    \item Calculate centroid of largest contour
    \item Map centroid position from -1 to 1 in image space and compute object velocity using time and calculated position from previous frame
    \item Calculate error from desired position in image space
    \item Calculate desired robot end-effector velocity in m/s from PD controller with \(k_p = 0.3\) and \(k_d = 0.06\). End-effector velocity in each axis is clamped to a maximum absolute speed of 300 mm/s.
    \item If hand is above workspace height, descend at 75 mm/s until workspace height is reached.
    \item If the normalised object position error satisfies \(|e_x| < 0.8\) and \(|e_y| < 0.15\) close the fingers (\(x\) is closing axis of fingers, \(y\) perpendicular).
\end{enumerate}

For the dual-camera perception system, a control output is calculated for each frame for both cameras (steps 1-6). The output from the camera with the larger measured object area in image space is actioned by the robot.

The system is implemented using a Franka-Emika Panda manipulator controlled through ROS. The wrist camera is an Intel RealSense D435i and the two hand camera units are constructed from a Raspberry Pi Camera Module V2 interfaced with a Raspberry Pi Zero.

\section{Results}
Fig. \ref{fig:GraspingResults} presents the success rate, average grasp time, and perception failure rate across the generated trajectories at workspace speeds between 100 and 200 mm/s. The grasp time was averaged across successful grasp attempts. The performance of the three perception systems proposed in section \ref{section:percetion_system} was evaluated.

The success rates for both cube sizes presented in Fig. \ref{fig:GraspingResults} demonstrate that the systems with camera configurations focused on the robots fingers out-perform the conventional wrist-mounted configuration on the dynamic grasping task. The difference is increasingly pronounced as the object speed increases. Similarly, the dual-hand camera system outperforms the single-hand camera solution.

The grasp time for each trial is dictated by the descent rate of the hand, and the time taken to close the fingers, which is independent of the perception system and object speed. Consequently there is no significant correlation between grasp time and perception system or object speed.

At low speeds, the success rate of all systems is high, and there are no failures due to loss of perception. At these speeds object motion can be observed and reacted to without the object leaving the cameras' field of view. As the object speed increases, the success rate of all perception systems decreases. In some failure cases the system maintained perception but collision with the fingers caused a failed grasp. Failures of this type could be reduced by refining the control system. However, failures resulting from sudden object motion that causes the object to exit the camera's field of view indicate the limitations of the perception systems. Failures of this type are represented by the perception failure rate in Fig. \ref{fig:GraspingResults}.

The results for the 30mm cube case presented in Fig. \ref{fig:30mmCubeResults} show a significant increase in perception failure rates for the wrist and single-hand systems with increasing object speed. The wrist-based system loses perception at a significantly higher rate than either the single- or dual-hand camera systems. By comparison, the system with dual-hand cameras maintained perception of the object in all cases at speeds up to 175 mm/s, and lost observation in only a single trial at the highest speed of 200 mm/s. The single-hand camera system demonstrated improved performance compared to the wrist-mounted configuration but lost perception in some cases at object speeds of 150 mm/s and greater.

The results for the 40mm cube presented in Fig. \ref{fig:40mmCubeResults} show that the single- and dual-hand camera systems experience the same loss of perception rates. The larger object is more likely to stay within the field of view, and therefore a single-hand camera is sufficient to maintain perception in this scenario. However, there is a slight improvement in grasp success rate for the dual-hand camera system even in the larger object case. This can likely be attributed to the quality of the perception data. With two cameras observing the object, when the object is at the limit of the field of view of one camera, the other is likely to have a clear view and can provide accurate feedback. 

\begin{figure}[t]
\vspace{10pt}
\centering
\includegraphics[width=\linewidth]{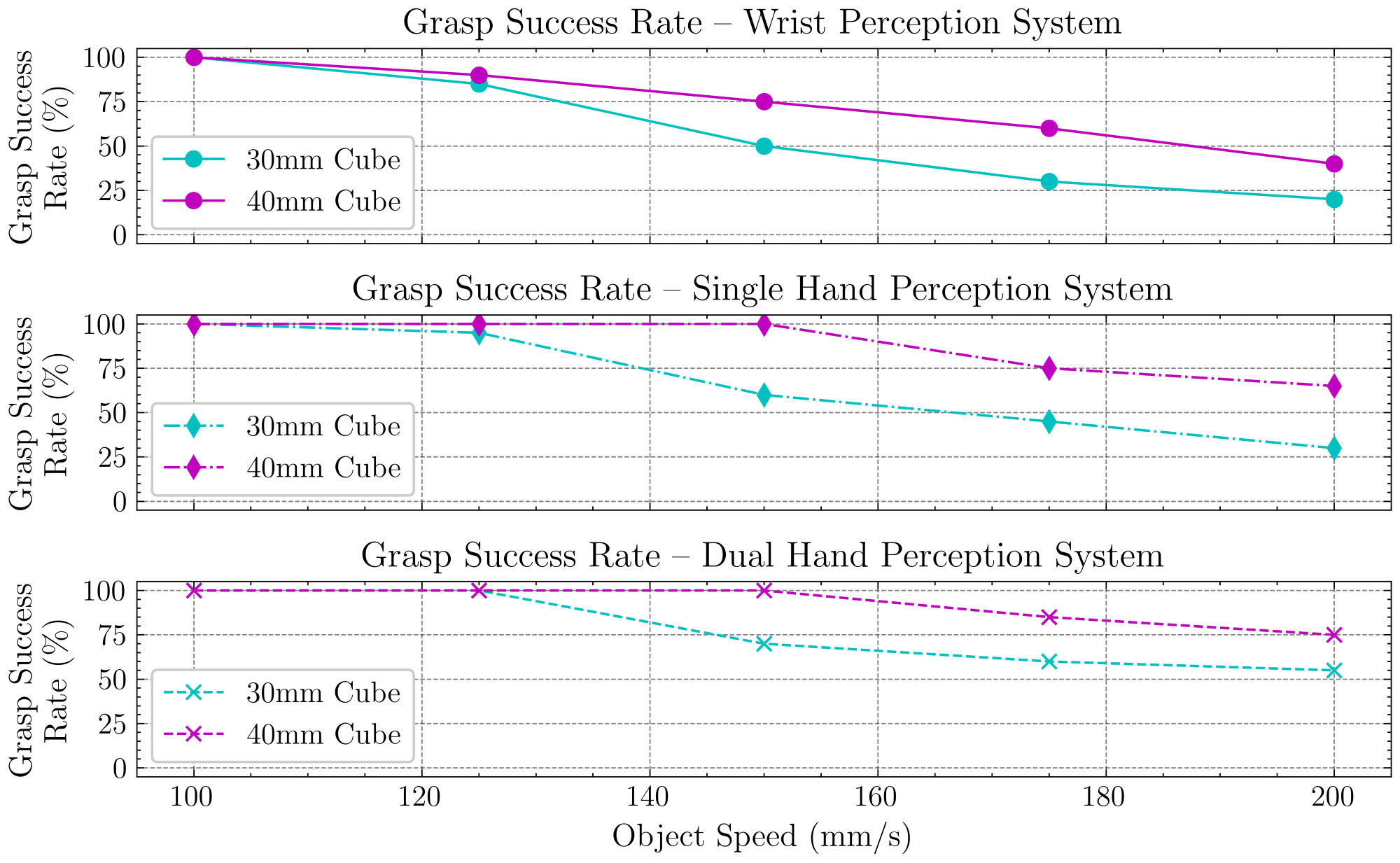}
\caption{Grasp success rates for perception systems on different cube sizes demonstrating improved performance on the larger object for all configurations.}
\label{fig:CubeSizeResults}
\end{figure}

Fig. \ref{fig:CubeSizeResults} presents a direct comparison of the grasp success rates on the two object sizes for each perception system. The larger object consistently results in an improved success rate. The increased size means that unexpected motion is less likely to completely move the object from the robot's fingers and requires less precise control once the object has been positioned between the fingers. Furthermore, the number of failures due to loss of perception is reduced because a larger object requires larger movements before it is completely removed from a camera's field of view. Although the difficulty of initial positioning of the gripper around an object increases with object size, the cubes are small relative to the initial gripper width (100 mm), so the effect is not significant. 

\addtolength{\textheight}{-0.6cm}

\section{Conclusion}
We have presented an open-sourced, reproducible benchmark for dynamic grasping and evaluated the performance of several perception system designs. DGBench was effective in enabling repeatable experiments for assessing grasping performance. The variability in speed and trajectory of the objects allowed for robust investigation of performance at varying levels of difficulty. The hardware is capable of recreating common dynamic grasping scenarios such as intermittent object motion, as well as predictable and unpredictable object trajectories. 

We proposed a multi-camera eye-in-hand grasping-perception system that demonstrated improved performance in a dynamic grasping task compared to a system with a conventional wrist-mounted camera. Our results show that considering the physical placement of perception systems is a crucial part of designing a robust robotics system. While this extends the design space and increases complexity, we see benefits particularly in highly dynamic environments. On experiments with a cube of size 30 mm and fast object motion of 200 mm/s, the proposed perception system achieved a grasp success rate of 55\% which is a 25\% improvement over the wrist-mounted camera baseline when using the same control law. Under these conditions, the baseline maintained accurate perception in only 20\% of trials, where our proposed system maintained observation in 95\% of cases. For more complex tasks, the proposed system should be paired with depth based sensing that can provide 3D information early in the grasp. 

Traditional camera placement for grasping perception systems has largely been a result of positioning sensors where it is convenient to do so. For static grasping tasks, the fixed, shoulder, and wrist mounting locations are suitable. However, grasping tasks requiring closed-loop feedback are increasingly common in mobile manipulation and similar dynamic environments. For these dynamic tasks, the challenges presented by occlusion, sensor minimum range, and sensor field of view highlight the deficiencies of conventional configurations. This paper demonstrates that camera placement is critical for developing reliable grasping perception systems. 

The proposed benchmark and perception system highlighted several opportunities for future investigation. The dynamic workspace hardware could be modified to include a $z$-axis rotation which is an important consideration when grasping more complex objects. The proposed approach to final phase perception could improve reliability in highly dynamic environments such as grasping from a mobile manipulator while in motion. Enabling closed-loop control through the entire grasp action could be used to implement robust grasping on low cost or soft robotic manipulators where imprecise control causes grasp failures.


\bibliographystyle{IEEEtran}
\bibliography{references}

\begin{thebibliography}{10}
\providecommand{\url}[1]{#1}
\csname url@samestyle\endcsname
\providecommand{\newblock}{\relax}
\providecommand{\bibinfo}[2]{#2}
\providecommand{\BIBentrySTDinterwordspacing}{\spaceskip=0pt\relax}
\providecommand{\BIBentryALTinterwordstretchfactor}{4}
\providecommand{\BIBentryALTinterwordspacing}{\spaceskip=\fontdimen2\font plus
\BIBentryALTinterwordstretchfactor\fontdimen3\font minus
  \fontdimen4\font\relax}
\providecommand{\BIBforeignlanguage}[2]{{%
\expandafter\ifx\csname l@#1\endcsname\relax
\typeout{** WARNING: IEEEtran.bst: No hyphenation pattern has been}%
\typeout{** loaded for the language `#1'. Using the pattern for}%
\typeout{** the default language instead.}%
\else
\language=\csname l@#1\endcsname
\fi
#2}}
\providecommand{\BIBdecl}{\relax}
\BIBdecl

\bibitem{MahlerBenchmark}
J.~Mahler, R.~Platt, A.~Rodriguez, M.~Ciocarlie, A.~Dollar, R.~Detry, M.~A.
  Roa, H.~Yanco, A.~Norton, J.~Falco, K.~v. Wyk, E.~Messina, J.~Leitner,
  D.~Morrison, M.~Mason, O.~Brock, L.~Odhner, A.~Kurenkov, M.~Matl, and
  K.~Goldberg, ``Guest editorial open discussion of robot grasping benchmarks,
  protocols, and metrics,'' \emph{IEEE Transactions on Automation Science and
  Engineering}, vol.~15, no.~4, pp. 1440--1442, 2018.

\bibitem{Kalashnikov}
D.~Kalashnikov, A.~Irpan, P.~Pastor, J.~Ibarz, A.~Herzog, E.~Jang, D.~Quillen,
  E.~Holly, M.~Kalakrishnan, V.~Vanhoucke, and S.~Levine, ``Scalable deep
  reinforcement learning for vision-based robotic manipulation,'' in
  \emph{Conference on Robot Learning}, 2018.

\bibitem{Allen}
P.~Allen, A.~Timcenko, B.~Yoshimi, and P.~Michelman, ``Automated tracking and
  grasping of a moving object with a robotic hand-eye system,'' \emph{IEEE
  Transactions on Robotics and Automation}, vol.~9, no.~2, pp. 152--165, 1993.

\bibitem{Akinola}
I.~Akinola, J.~Xu, S.~Song, and P.~K. Allen, ``Dynamic grasping with
  reachability and motion awareness,'' in \emph{2021 IEEE/RSJ International
  Conference on Intelligent Robots and Systems}.\hskip 1em plus 0.5em minus
  0.4em\relax IEEE, 2021, pp. 9422--9429.

\bibitem{Viereck}
U.~Viereck, A.~T. Pas, K.~Saenko, and R.~W. Platt, ``Learning a visuomotor
  controller for real world robotic grasping using simulated depth images,'' in
  \emph{Conference on Robotic Learning}, 2017, pp. 291--300.

\bibitem{Tuscher}
M.~Tuscher, J.~H{\"o}rz, D.~Driess, and M.~Toussaint, ``Deep 6-dof tracking of
  unknown objects for reactive grasping,'' in \emph{2021 IEEE International
  Conference on Robotics and Automation}.\hskip 1em plus 0.5em minus
  0.4em\relax IEEE, 2021, pp. 14\,185--14\,191.

\bibitem{MorrisonLearningRobust}
D.~Morrison, P.~Corke, and J.~Leitner, ``Learning robust, real-time, reactive
  robotic grasping,'' \emph{The International Journal of Robotics Research},
  vol.~39, no. 2-3, pp. 183--201, 2020.

\bibitem{Song}
S.~Song, A.~Zeng, J.~Lee, and T.~Funkhouser, ``Grasping in the wild: Learning
  6dof closed-loop grasping from low-cost demonstrations,'' \emph{Robotics and
  Automation Letters}, 2020.

\bibitem{Bonsignorio}
F.~Bonsignorio, ``A new kind of article for reproducible research in
  intelligent robotics [from the field],'' \emph{IEEE Robotics Automation
  Magazine}, vol.~24, no.~3, pp. 178--182, 2017.

\bibitem{Calli}
B.~Calli, A.~Walsman, A.~Singh, S.~Srinivasa, P.~Abbeel, and A.~M. Dollar,
  ``Benchmarking in manipulation research: Using the yale-cmu-berkeley object
  and model set,'' \emph{IEEE Robotics Automation Magazine}, vol.~22, no.~3,
  pp. 36--52, 2015.

\bibitem{Shimoga}
K.~Shimoga, ``Robot grasp synthesis algorithms: A survey,'' \emph{The
  International Journal of Robotics Research}, vol.~15, no.~3, pp. 230--266,
  1996.

\bibitem{Bicchi}
A.~{Bicchi} and V.~{Kumar}, ``Robotic grasping and contact: a review,'' in
  \emph{Proceedings IEEE International Conference on Robotics and Automation.},
  vol.~1, 2000, pp. 348--353.

\bibitem{Sahbani}
A.~Sahbani, S.~El-Khoury, and P.~Bidaud, ``An overview of 3d object grasp
  synthesis algorithms,'' \emph{Robotics and Autonomous Systems}, vol.~60,
  no.~3, pp. 326--336, 2012.

\bibitem{Bohg}
J.~{Bohg}, A.~{Morales}, T.~{Asfour}, and D.~{Kragic}, ``Data-driven grasp
  synthesis—a survey,'' \emph{IEEE Transactions on Robotics}, vol.~30, no.~2,
  pp. 289--309, 2014.

\bibitem{Kleeberger}
K.~Kleeberger, R.~Bormann, W.~Kraus, and M.~Huber, ``A survey on learning-based
  robotic grasping,'' \emph{Current Robotics Reports}, vol.~1, p. 239–249, 12
  2020.

\bibitem{Chebotar}
Y.~Chebotar, K.~Hausman, O.~Kroemer, G.~S. Sukhatme, and S.~Schaal,
  ``Generalizing regrasping with supervised policy learning,'' in \emph{2016
  International Symposium on Experimental Robotics}, D.~Kuli{\'{c}},
  Y.~Nakamura, O.~Khatib, and G.~Venture, Eds.\hskip 1em plus 0.5em minus
  0.4em\relax Cham: Springer International Publishing, 2017, pp. 622--632.

\bibitem{Yu}
Y.~Yu, Z.~Cao, S.~Liang, W.~Geng, and J.~Yu, ``A novel vision-based grasping
  method under occlusion for manipulating robotic system,'' \emph{IEEE Sensors
  Journal}, vol.~20, no.~18, pp. 10\,996--11\,006, 2020.

\bibitem{Calandra}
R.~Calandra, A.~Owens, D.~Jayaraman, J.~Lin, W.~Yuan, J.~Malik, E.~H. Adelson,
  and S.~Levine, ``More than a feeling: Learning to grasp and regrasp using
  vision and touch,'' \emph{IEEE Robotics and Automation Letters}, vol.~3,
  no.~4, p. 3300–3307, Oct 2018.

\bibitem{Farias}
C.~de~Farias, N.~Marturi, R.~Stolkin, and Y.~Bekiroglu, ``Simultaneous tactile
  exploration and grasp refinement for unknown objects,'' \emph{IEEE Robotics
  and Automation Letters}, vol.~6, no.~2, p. 3349–3356, Apr 2021.

\bibitem{Hutchinson}
S.~Hutchinson, G.~Hager, and P.~Corke, ``A tutorial on visual servo control,''
  \emph{IEEE Transactions on Robotics and Automation}, vol.~12, no.~5, pp.
  651--670, 1996.

\bibitem{Kamon}
I.~{Kamon}, T.~{Flash}, and S.~{Edelman}, ``Learning to grasp using visual
  information,'' in \emph{Proceedings of IEEE International Conference on
  Robotics and Automation}, vol.~3, 1996, pp. 2470--2476 vol.3.

\bibitem{Lin}
C.~Lin, Y.-L. Chen, W.~Hao, and X.~Wu, ``Occluded object grasping based on
  robot stereo vision,'' in \emph{Proceedings of the 10th World Congress on
  Intelligent Control and Automation}, 2012, pp. 3698--3704.

\bibitem{Marturi}
N.~Marturi, M.~Kopicki, A.~Rastegarpanah, V.~Rajasekaran, M.~Adjigble,
  R.~Stolkin, A.~Leonardis, and Y.~Bekiroglu, ``Dynamic grasp and trajectory
  planning for moving objects,'' \emph{Autonomous Robots}, vol.~43, pp.
  1241--1256, 2019.

\bibitem{Pinto}
L.~Pinto and A.~Gupta, ``Supersizing self-supervision: Learning to grasp from
  50k tries and 700 robot hours,'' \emph{IEEE International Conference on
  Robotics and Automation}, pp. 3406--3413, 2016.

\bibitem{Levine}
S.~Levine, P.~Pastor, A.~Krizhevsky, J.~Ibarz, and D.~Quillen, ``Learning
  hand-eye coordination for robotic grasping with deep learning and large-scale
  data collection,'' \emph{The International Journal of Robotics Research},
  vol.~37, no. 4-5, pp. 421--436, 2018.

\bibitem{Logothetis}
M.~Logothetis, G.~C. Karras, S.~Heshmati-Alamdari, P.~Vlantis, and K.~J.
  Kyriakopoulos, ``A model predictive control approach for vision-based object
  grasping via mobile manipulator,'' in \emph{2018 IEEE/RSJ International
  Conference on Intelligent Robots and Systems}.\hskip 1em plus 0.5em minus
  0.4em\relax IEEE, 2018, pp. 1--6.

\bibitem{Kahn}
G.~Kahn, P.~Sujan, S.~Patil, S.~Bopardikar, J.~Ryde, K.~Goldberg, and
  P.~Abbeel, ``Active exploration using trajectory optimization for robotic
  grasping in the presence of occlusions,'' in \emph{IEEE International
  Conference on Robotics and Automation}, 2015, pp. 4783--4790.

\bibitem{MorrisonMultiView}
D.~Morrison, P.~Corke, and J.~Leitner, ``Multi-view picking: Next-best-view
  reaching for improved grasping in clutter,'' in \emph{International
  Conference on Robotics and Automation}, 2019, pp. 8762--8768.

\bibitem{Mousavian}
A.~Mousavian, C.~Eppner, and D.~Fox, ``6-dof graspnet: Variational grasp
  generation for object manipulation,'' \emph{IEEE/CVF International Conference
  on Computer Vision}, pp. 2901--2910, 2019.

\bibitem{ZengTossingBot}
A.~Zeng, S.~Song, J.~Lee, A.~Rodriguez, and T.~Funkhouser, ``Tossingbot:
  Learning to throw arbitrary objects with residual physics,'' \emph{IEEE
  Transactions on Robotics}, vol.~36, no.~4, pp. 1307--1319, 2020.

\bibitem{Haviland}
J.~Haviland, F.~Dayoub, and P.~Corke, ``Control of the final-phase of
  closed-loop visual grasping using image-based visual servoing,'' \emph{arXiv
  preprint arXiv:2001.05650}, 2020.

\bibitem{Berscheid}
L.~Berscheid, P.~Meißner, and T.~Kröger, ``Robot learning of shifting objects
  for grasping in cluttered environments,'' in \emph{IEEE/RSJ International
  Conference on Intelligent Robots and Systems}, 2019.

\bibitem{Johns}
E.~Johns, S.~Leutenegger, and A.~J. Davison, ``Deep learning a grasp function
  for grasping under gripper pose uncertainty,'' in \emph{IEEE/RSJ
  International Conference on Intelligent Robots and Systems}, 2016, pp.
  4461--4468.

\bibitem{Saxena}
A.~Saxena, J.~Driemeyer, and A.~Y. Ng, ``Robotic grasping of novel objects
  using vision,'' \emph{The International Journal of Robotics Research},
  vol.~27, no.~2, pp. 157--173, 2008.

\bibitem{Arora}
P.~Arora and C.~Papachristos, ``Mobile manipulator robot visual servoing and
  guidance for dynamic target grasping,'' in \emph{Advances in Visual
  Computing}, G.~Bebis, Z.~Yin, E.~Kim, J.~Bender, K.~Subr, B.~C. Kwon,
  J.~Zhao, D.~Kalkofen, and G.~Baciu, Eds.\hskip 1em plus 0.5em minus
  0.4em\relax Cham: Springer International Publishing, 2020, pp. 223--235.

\bibitem{Zimmermann}
S.~Zimmermann, R.~Poranne, and S.~Coros, ``Go fetch!-dynamic grasps using
  {B}oston {D}ynamics {S}pot with external robotic arm,'' in \emph{2021 IEEE
  International Conference on Robotics and Automation}.\hskip 1em plus 0.5em
  minus 0.4em\relax IEEE, 2021, pp. 4488--4494.

\end{thebibliography}

\end{document}